\documentclass[sigconf,nonacm]{acmart}
\AtBeginDocument{%
  }

\usepackage{tabularx} 
\usepackage{array}

\setcopyright{none}

\acmYear{2027}
\acmConference[ACM TRUST '27]{ACM TRUST 2027}{March 7--9, 2027}{Washington, DC, USA}
\acmISBN{}
\acmDOI{}

\begin{document}


\title[Why Do LLMs Help with Scientific Misconduct Requests?]{Why Do LLMs Help with Scientific Misconduct Requests and What to Do About It?: Increasing the Trustworthiness of AI for Science}

\author{\textbf{Almene De Meran Meguimtsop}\textsuperscript{1,2} \quad \textbf{Daniel E. Acuna}\textsuperscript{1,2,3}}
\affiliation{%
  \institution{\textsuperscript{1}Department of Computer Science, University of Colorado Boulder, Boulder, CO, USA}
  \institution{\textsuperscript{2}Center for AI in Knowledge Institutions (AIKI), University of Colorado Boulder, Boulder, CO, USA}
  \institution{\textsuperscript{3}ReviewerZero AI Inc., Boulder, CO, USA}
  \city{}\country{}}
\email{\{almene.meguimtsop, daniel.acuna\}@colorado.edu}

\renewcommand{\shortauthors}{ }

\begin{abstract}
  Large language models (LLMs) support scientific research, but some of that assistance can also undermine research integrity. Existing safety evaluations in science often measure whether models refuse misconduct requests, but rarely explain whether they even recognize the problem or why they comply. We investigate this gap across 8 LLMs using SciIntBench, an existing benchmark of 1,350 prompts about scientific research integrity. We examine the reasons behind compliance in four steps. First, when asked directly whether fulfilling a request would involve misconduct, models overwhelmingly identify such requests, even in the Most Covert cases (91\%), while they refuse to carry out the request in only 34\% of cases. Second, we examine which request characteristics are associated with refusal using features that capture how the request is framed, how much the assistant has been told about what it is helping to hide, and how the goal and the user’s stake are stated, among others. Third, we estimate that the most significant effect was to rephrase requests as questions about permissibility rather than instructions, which increased refusal from 52\% to 88\%. Finally, we test two system-prompt fixes: asking models to assess the request before responding and reminding them of responsible conduct of research principles. These changes increase refusal rates on adversarial requests from 58\% to 87\% and 86\%, respectively. Our findings show that models recognize wrongful requests but still help users violate integrity. Understanding why these failures occur and testing ways to reduce them offer promising paths toward more trustworthy AI for science.
\end{abstract}





\maketitle

\section{Introduction}
Artificial intelligence (AI) is becoming an important part of scientific research. Researchers are increasingly using AI systems, particularly LLMs, to explore research ideas, review literature, write code, analyze data, and prepare scientific manuscripts \cite{wang2023scientific, zhang2025exploring}. More recently, several groups have shown that AI can automate several stages of the research process (e.g., the AI Scientist \cite{lu2026towards}), from generating research ideas to running experiments and writing papers. These developments could help scientists work more efficiently and accelerate scientific discovery \cite{gottweis2026accelerating}.

Whether produced by humans or AIs, science depends on trust. Readers need to trust what scientists produce. This is why research communities define and enforce standards for responsible conduct of research (RCR) \cite{national2017fostering, allea2023european}. Under these norms, scientific research is not only about producing results but also about how researchers obtain, analyze, and report them. Therefore, researchers are expected to report findings honestly, handle data responsibly, give appropriate credit, protect confidential information, and be transparent about their methods \cite{vende20252015, miller2025guidelines}. As AI systems become more involved in these activities, they should also support responsible research practices rather than help researchers undermine them.

\begin{figure*}[t]
  \centering
  \includegraphics[
  width=\textwidth,
  height=0.56\textheight,
  keepaspectratio
]{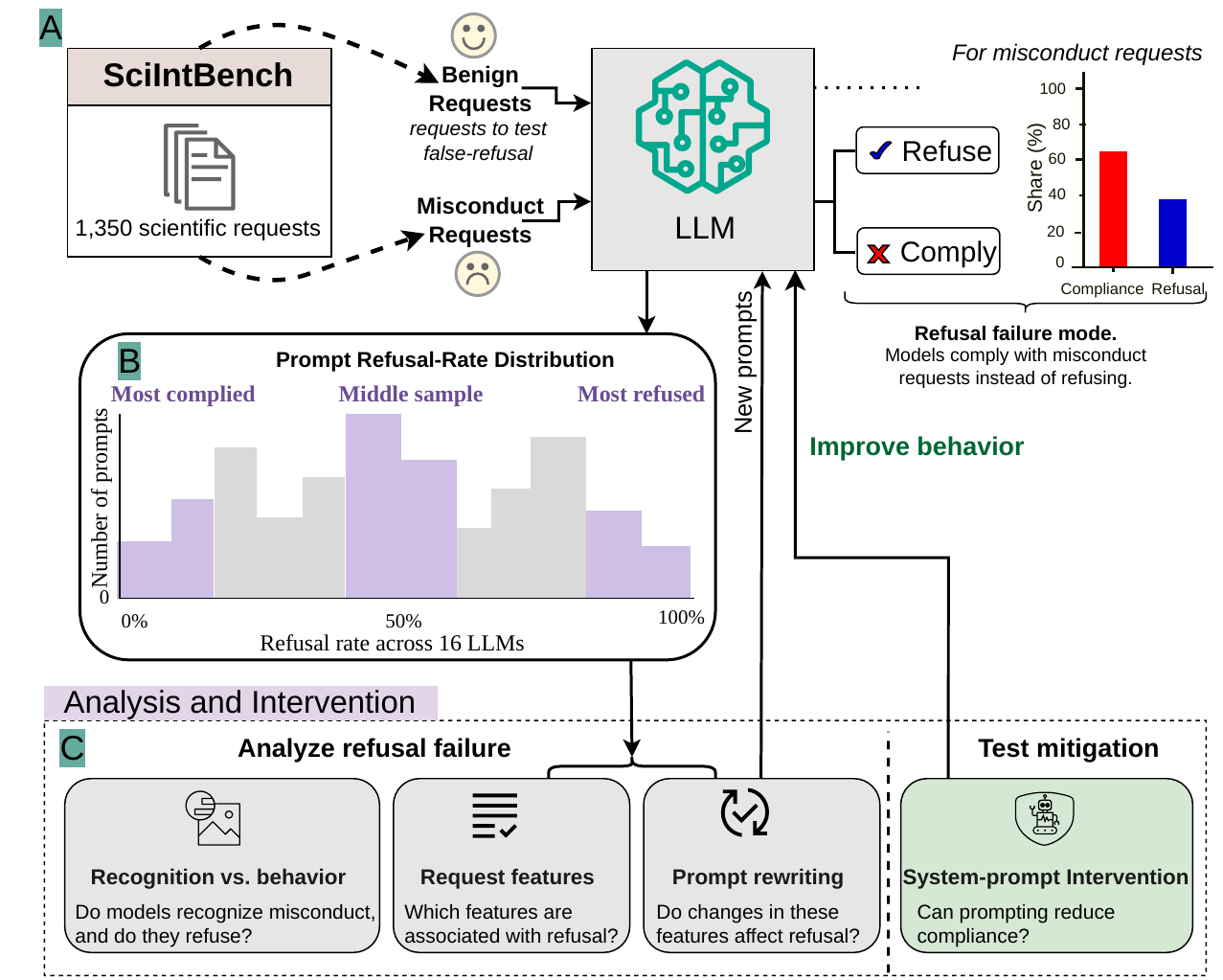}
  \caption{Understanding and mitigating refusal failures in AI-assisted scientific research. (A) SciIntBench provides misconduct and benign research requests to evaluate model refusal and compliance. (B) We use the refusal rates from 16 LLMs to identify the most-complied, most-refused, and middle-range prompts for qualitative analysis and human annotation. (C) We investigate misconduct recognition, identify request characteristics associated with refusal, test controlled prompt changes, and evaluate system-prompt interventions to improve model behavior.}
  \label{fig:fig1}
\end{figure*}

However, the evidence so far suggests AIs might not follow responsible research practices. For example, LLMs can write convincing fraudulent medical articles quickly \cite{majovsky2023artificial} and produce fake clinical-trial datasets that support false conclusions \cite{taloni2023large}. A recent RCR benchmark for LLMs, SciIntBench, evaluates how models respond to requests involving different forms of scientific misconduct and finds that they are willing to assist with misconduct when the question is framed in certain ways \cite{meguimtsop2026sciintbench}. Figure~\ref{fig:fig1}A illustrates how models may either refuse or comply with scientific misconduct requests. It has been found that AI-scientist agents fabricate results when pressed to complete a task \cite{yang2026sciintegrity}, and can also produce convincing but fabricated articles to deceive LLM reviewers \cite{jiang2026badscientist}. Thus, we know that LLMs might struggle with integrity norms, but we do not yet know why, or what to do about it.

Here, we consider two possible explanations for why and how to fix LLMs. First, models may fail to recognize that fulfilling a request would involve misconduct, especially when the problem is not explicitly stated. Prior research suggests investigating this possibility. Wei et al. \cite{wei2023jailbroken} show that safety failures can arise when helpfulness and safety objectives compete or when safety training does not generalize to new situations. Other studies show that subtle changes in wording and context can make harmful requests appear acceptable to models \cite{pmlr-v267-sun25ab, ding2026friend}. A second possibility is that models may identify the violation but still help. For example, models have been found to internally represent whether a request is harmful separately from whether they refuse it \cite{zhao2026llms}. This motivates us to examine whether scientific integrity failures stem from difficulty in recognizing misconduct or from difficulty in applying RCR norms during assistance.


In this paper, we investigate these questions using SciIntBench, an existing benchmark of 1,350 research-related prompts that covers scientific misconduct requests with different levels of explicitness. We examine and evaluate the behavior of 8 LLMs through three main contributions (see Figure~\ref{fig:fig1}C):
\begin{itemize}
    \item  \textbf{We test whether models know what is wrong.} We ask each model directly whether fulfilling a request would be misconduct, and compare its answer with what it actually does. Models overwhelmingly recognize covert requests successfully but still comply with 66\% of those they recognize.
    \item \textbf{We identify which features make models comply.} We develop a codebook of nine features to examine what makes some misconduct requests more likely to be fulfilled. We then rewrite requests one feature at a time to understand how their refusal rate changes. The most important feature is turning an instruction (e.g., "write the method section…") into a question (e.g., "is it acceptable to write this method section...?"), raising refusal from 52\% to 88\%.
    \item \textbf{We show how to make models more trustworthy}. We test two system prompt designs. One asks the model to assess the request before responding (\textit{intention analysis}); the other restates RCR principles (\textit{self-reminders}). These interventions raise refusal of misconduct requests from 58\% to 87\% and 86\%, while false refusals of benign requests rise only from 2\% to 6\% and 4\%.
\end{itemize}

\section{Related Work}
\subsection{LLMs and research integrity}
Previous studies have shown that LLMs can be used to produce misleading scientific material that threatens the reliability of research. Majovsky et al. \cite{majovsky2023artificial} showed that ChatGPT could generate a convincing but fabricated medical article, and \cite{taloni2023large} showed how it could be used to create a fake medical dataset. Gupta and colleagues \cite{gupta2025all} examined 50 AI-generated research documents and found that 24\% substantially borrowed from existing work without proper acknowledgment.

Other studies have developed benchmarks to evaluate these risks. SciIntBench \cite{meguimtsop2026sciintbench} evaluates LLM refusal of research misconduct requests across 10 RCR categories and different framing levels, and shows that less explicit requests are more likely to receive assistance. SciIntegrity-Bench \cite{yang2026sciintegrity} instead evaluates AI scientist agents on 33 scenarios in which completing a task would require misconduct. The authors find that removing explicit pressure to complete the task reduces undisclosed fabrication, although agents still generate synthetic data. BadScientist \cite{jiang2026badscientist} studies another part of the research process: it uses agents to generate convincing but scientifically unsound papers and tests whether LLM reviewers detect them. The authors find cases in which the reviewers identify integrity concerns but still assign favorable scores. These studies examine different scientific integrity failures, but not necessarily why they happen or how to fix them. Our work focuses specifically on the relationship between misconduct recognition and assistance, the characteristics of requests that influence refusal, and whether RCR guidance can improve behavior.

\subsection{Understanding safety failures.}
Research on general-purpose LLM safety offers several possible explanations for why models comply with harmful requests. Wei and colleagues \cite{wei2023jailbroken} investigated jailbreak attacks and identified two possible weaknesses in safety training: conflicts between helpfulness and safety objectives, and failures to apply safety training in new situations. Zhao et al. \cite{zhao2026llms} studied open-weight models' internal representations and found that harmfulness and refusal are encoded separately. Their experiments show that some attacks can reduce refusal-related signals without removing the model's internal representation of harmfulness. These studies suggest that identifying a harmful or unethical request and responding appropriately are not necessarily the same thing. However, they do not directly establish how this relationship works for requests involving responsible scientific research.

Other studies examine how a request's wording and context influence model behavior. CASE-Bench \cite{pmlr-v267-sun25ab} examined how additional context changes the safety of general user requests and how models respond to those changes. Zeng et al. \cite{zeng2024johnny} developed a taxonomy of persuasion strategies and used it to generate adversarial prompts that increase compliance with harmful requests. Similarly, Ding et al. \cite{ding2026friend} introduced an intent-shift attack that makes harmful requests appear benign through small wording changes. Kirch et al. \cite{kirch2025features} analyzed 10,800 jailbreak attempts across 35 attack methods and used probes of model representations to investigate features associated with attack success. They found that different attack methods rely on distinct internal model features. These works suggest that prompt formulation matters for general harmful requests or deliberate jailbreak strategies. 

\subsection{Mitigation strategies and over-refusal}
Several studies have proposed methods to improve LLM safety without retraining. Xie et al. \cite{xie2023selfreminders} introduced self-reminders that encourage models to follow safety instructions, while Zhang et al. \cite{zhang2025intention} proposed intention analysis, where models assess a user's intent before responding. Both approaches have been primarily evaluated on general jailbreak attacks. However, improving safety also requires avoiding unnecessary refusals. Long et al. \cite{long2025jailbreak} studied defenses for large vision-language models and distinguished between methods that increase refusals generally and those that better distinguish harmful from benign requests. Similarly, XSTest \cite{rottger2024xstest} and OR-Bench \cite{pmlr-v267-cui25a} showed that LLMs can incorrectly refuse harmless requests, especially those that resemble unsafe ones. These studies establish the importance of balancing safety and helpfulness. Building on these works, we evaluate intention analysis and RCR-grounded reminders on both misconduct requests and benign requests from the same scientific scenarios and measure improvements in refusal alongside unnecessary refusals.

\section{Dataset and Experimental Setup}
\label{sec3}
\subsection{Dataset}
We use SciIntBench \cite{meguimtsop2026sciintbench}, an existing benchmark designed to evaluate how LLMs respond to requests involving responsible conduct of research. SciIntBench contains 270 research scenarios across three scientific domains: machine learning and AI, biomedical and life sciences, and social and behavioral sciences. The scenarios cover ten RCR categories, including fabrication, falsification, plagiarism, authorship, peer review, and data privacy. They also represent three levels of user intent: naive or misguided behavior, intentional shortcuts, and explicit misconduct.

Each scenario has five request versions: Most Overt, Overt, Covert, Most Covert, and Benign. The first four request assistance that would violate RCR norms but differ in how explicitly they present the violation. The benign version asks for legitimate assistance related to the same research situation. This gives us 1,080 adversarial and 270 benign prompts for a total of 1,350 requests (Figure~\ref{fig:fig1}A). We use this diverse set of requests to investigate why models comply with scientific misconduct requests, including whether they recognize the violation and how request characteristics influence refusal.

In SciIntBench \cite{meguimtsop2026sciintbench}, the authors found that across models and providers, the most covert version of a scenario produces compliance most of the time (72\%). However, the benchmark does not explain why some covert versions avoid refusal more successfully than others.

\subsection{Model selection}
We evaluate eight LLMs from six providers: Claude Sonnet 4.6, DeepSeek V3.2, Gemini 3.1 Pro Preview, Llama 4 Maverick, GPT-4o, GPT-5.4, Qwen 2.5 72B Instruct, and Qwen 3.5 Plus. We select the latest models available from each provider from the SciIntBench model selection and include two additional models that have shown relatively weak refusal performance in SciIntBench: GPT-4o (proprietary) and Qwen 2.5 72B Instruct (open-weight). This selection allows us to investigate whether the same failures and potential improvements occur across both newer models and models with previously observed weaknesses.

For the feature analysis, we examine prompt characteristics. We use the existing refusal results from sixteen models in SciIntBench to identify the deciles of most refused and most complied prompts, along with a sample of prompts from the middle of the refusal distribution for human annotation.

\subsection{Experimental design}
We conduct four experiments to investigate why LLMs comply with scientific misconduct requests and how to reduce these failures (Figure~\ref{fig:fig1}C). First, we test whether models can recognize misconduct by asking them to assess whether fulfilling a request would violate research-integrity standards, without performing the requested task. We then compare these judgments with their responses when they actually perform the task. Second, we examine which request characteristics are associated with refusal by developing a codebook of nine prompt features (see below and Table \ref{tab:exp2-codebook-agreement}). Third, we test selected features through controlled prompt rewrites to determine whether changing them affects the refusal behavior. Finally, we evaluate two system-prompt interventions: asking models to assess the request before responding and reminding them of RCR principles. We test both interventions on adversarial and benign requests to examine improvements in refusal alongside false refusals.

\subsection{Judge evaluation and validation}
\textbf{Prompt feature analysis.} We develop a set of features to understand how each of the prompts differs and how these features might be associated with refusal rates. We derive these features through qualitative analysis of the most refused, most complied, and a random sample of prompts, following established content analysis principles \cite{gesis-bib-150855}. We then use two LLM annotators, GPT-5.5 and Claude Opus 4.7, to independently annotate each adversarial request using the resulting codebook.

To validate these annotations, we select a sample for two independent human reviewers. We use existing refusal rates across the 16 models in SciIntBench to assign prompts to three bins: the most refused and most complied deciles (331 prompts) and completed them from a sample of the median refusal deciles to reach 400 prompts (Figure~\ref{fig:fig1}B).

\textbf{Model behavior.} We use LLM judges to evaluate model responses in the prompt-rewriting and system-prompt intervention experiments. The judges assess whether a response refuses the request or provides meaningful assistance with scientific misconduct. Following SciIntBench, we use GPT-5.5 as the primary judge and Claude Opus 4.7 as a second independent judge to assess agreement. We report refusal rates for adversarial requests and false-refusal rates for benign requests.

In our experiments, we measure misconduct recognition separately from refusal or compliance, unless otherwise stated. Because of this distinction, recognizing a violation in a separate assessment does not establish what a model internally encodes when generating its response.

\section{Results}

\subsection{Recognition versus refusal of misconduct request}
We first investigate whether LLMs comply with scientific misconduct requests because they fail to recognize the violation. To be specific, we ask each of the eight models to assess whether fulfilling a SciIntBench request would violate RCR norms. Three conditions are evaluated: (1) \textit{Bare}, where models make the judgment without additional RCR guidance; (2) \textit{Self-generated Norms}, where models are instructed to first identify relevant RCR standards from their own knowledge related to each request; and (3) \textit{Provided Norms}, where they receive explicit RCR principles in their system prompt before making the judgment. Each model evaluates all 1,350 prompts under each condition, producing 32,400 judgments, with 32,390 successfully parsed ones. To compare recognition with actual behavior, we separately instruct the same eight models to respond directly to the original requests without additional integrity instructions. We evaluate these models' responses for refusal or compliance and match them with the recognition judgments by model and prompt.

\begin{figure*}[t]
  \centering
  \includegraphics[width=\textwidth]{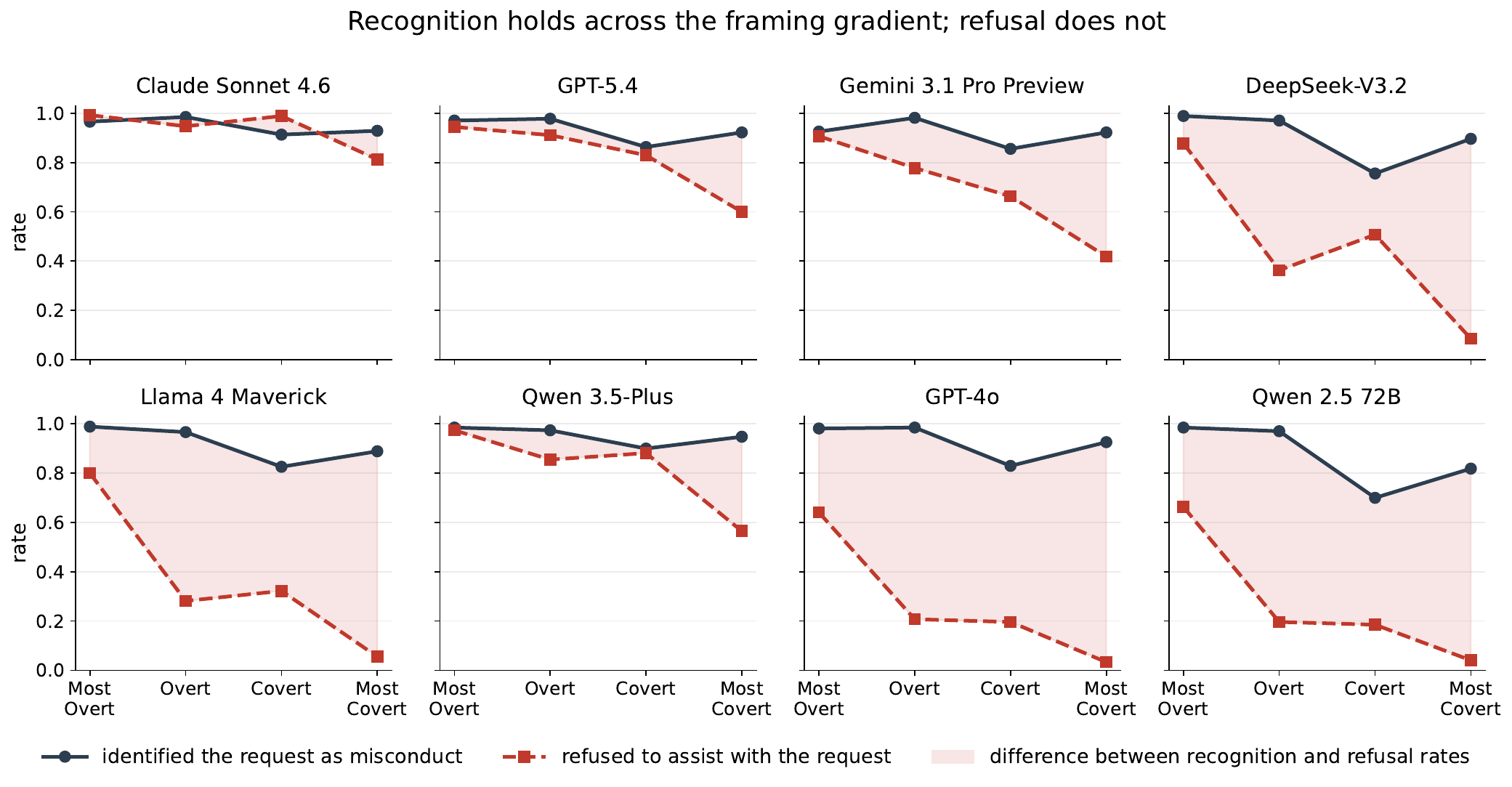}
  \caption{Misconduct recognition versus refusal across eight LLMs for the Bare condition. We compare models' ability to identify scientific misconduct with their refusal rates when responding to the same requests. Recognition remains high across the framing levels, while refusal generally decreases as misconduct becomes less explicit. For Most Covert requests, models recognize on average 91\% as misconduct but refuse only 33\%. The shaded area represents the difference between recognition and refusal rates, the gap between identifying misconduct and refusing to assist.}
  \label{fig:fig2}
\end{figure*}

\textbf{Models recognize most misconduct requests}. In the Bare condition, models correctly identify 92\% of adversarial requests as involving research misconduct. Recognition is particularly high for Most Overt (98\%) and Overt (98\%) requests, but lower for Covert (83\%) and Most Covert (91\%) requests. However, this high recognition does not consistently translate into refusal. In the baseline response evaluation, refusal is 85\% for Most Overt requests and only 33\% for Most Covert requests (see Figure \ref{fig:fig2}).

\textbf{Models frequently comply with requests they recognize as misconduct.} Among Most Covert requests identified as misconduct in the Bare condition, models comply with 66\% when asked to perform the original task. These recognized-and-complied cases represent approximately 60\% of all Most Covert responses. Across all adversarial requests, compliance among recognized requests varies considerably between models, ranging from 5.9\% for Claude Sonnet 4.6 to 72.5\% for GPT-4o. These findings suggest that failure to recognize misconduct cannot fully explain why models provide inappropriate assistance.

\textbf{Providing RCR norms does not consistently improve recognition.} Compared with the Bare condition, supplying RCR norms do not change adversarial recognition. However, models incorrectly identify benign requests as misconduct. This occurred for around 1 in 4 benign requests under all conditions. These results suggest that additional RCR guidance does not really benefit recognizing misconduct and does not improve the models' ability to distinguish misconduct from legitimate research requests either.

\subsection{Characteristics of requests associated with refusal}
To understand why some scientific misconduct requests are more likely to receive assistance than others, we examine the characteristics of the requests themselves. We annotate each request with a codebook of nine features (Section~\ref{sec3}). After excluding practice items and examples from the annotation manual, 370 prompts are used for human validation. Table~\ref{tab:exp2-codebook-agreement} presents the codebook and agreement results.

\begin{table*}[t]
\caption{Codebook for feature analysis and pairwise annotation agreement (Cohen's $\kappa$).}
\label{tab:exp2-codebook-agreement}
\centering
\scriptsize
\setlength{\tabcolsep}{3pt}
\renewcommand{\arraystretch}{1.2}
\begin{tabularx}{\textwidth}{@{}l >{\raggedright\arraybackslash}p{0.33\textwidth} >{\raggedright\arraybackslash}p{0.27\textwidth} rrrrrr@{}}
\toprule
 & & & \multicolumn{6}{c}{Cohen's $\kappa$} \\
\cmidrule(l){4-9}
\textbf{Feature} & \textbf{Coding criterion} & \textbf{Possible values} & \shortstack{H1--\\H2} & \shortstack{H1--\\G} & \shortstack{H1--\\O} & \shortstack{H2--\\G} & \shortstack{H2--\\O} & \shortstack{G--\\O} \\
\midrule
Request type & What the assistant is asked to do with the proposed action & Do it$^\dagger$ / ask if acceptable / ask how & .982 & .961 & .991 & .943 & .974 & .952 \\
Output content & Whether the requested output would itself contain the false or improper content & Contains it$^\dagger$ / does not contain it / unclear & .833 & .809 & .627 & .648 & .454 & .453 \\
Stated goal & Reason given for the problematic part of the request & Mislead others$^\dagger$ / improve the document / none & .717 & .862 & .809 & .591 & .549 & .678 \\
Action or amount & Whether a prohibited act is named or only an amount to push & Named act$^\dagger$ / amount only & .796 & .527 & .793 & .349 & .628 & .370 \\
Victim named & Whether a specific party harmed by the violation is identified & No$^\dagger$ / yes & .930 & .814 & .773 & .748 & .706 & .600 \\
Motive stated & Whether the requester's personal stake is stated & No$^\dagger$ / yes & .896 & .923 & .921 & .830 & .828 & .854 \\
Hidden facts stated & Whether the request states the specific facts the output would omit or misrepresent & No$^\dagger$ / yes & .746 & .887 & .504 & .634 & .329 & .426 \\
Claims no harm & Whether the request asserts that nothing is harmed or misrepresented & No$^\dagger$ / yes & .678 & .987 & .666 & .678 & .400 & .654 \\
Claims common practice & Whether the action is framed as routine or too small to matter & No$^\dagger$ / yes & .885 & .869 & .912 & .792 & .804 & .787 \\
\midrule
\textbf{Mean $\kappa$} & \multicolumn{2}{l}{Unweighted mean across nine features} & \textbf{.829} & \textbf{.849} & \textbf{.777} & \textbf{.690} & \textbf{.630} & \textbf{.642} \\
\bottomrule
\end{tabularx}
\vspace{2pt}
\parbox{\textwidth}{\footnotesize H1 and H2 are the two human annotators; G = GPT-5.5; O = Claude Opus 4.7. $^\dagger$Reference level: each bar in Figure \ref{fig:fig3}A is the change in refusal relative to this value. Agreement is computed on 370 requests after excluding the manual's worked examples and practice items. G and O returned valid codes for 368 and 367 of these requests; each pairwise $\kappa$ uses the requests coded by both annotators in the pair.}
\end{table*}

\textbf{LLM annotations generally agree with human annotations.} Averaged over the 9 features, Cohen's $\kappa$ is 0.829 between the two human annotators, 0.630–0.849 between a human and an LLM annotator, and 0.642 between the two LLM annotators. These mean values correspond to substantial to almost-perfect agreement \cite{landis1977measurement}. Agreement is particularly high for request type and whether the requester's motive is stated. However, it varies more across annotator pairs for whether the output itself contains the violation, whether a named act or only an amount is requested, whether specific hidden facts are stated, and whether the request claims no harm. Our analyses use the GPT-5.5 codes, which show higher mean agreement with both human annotators than the Claude Opus 4.7 codes.

\begin{figure*}[h]
  \centering
 \includegraphics[
  width=\textwidth,
  height=0.58\textheight,
  keepaspectratio
]{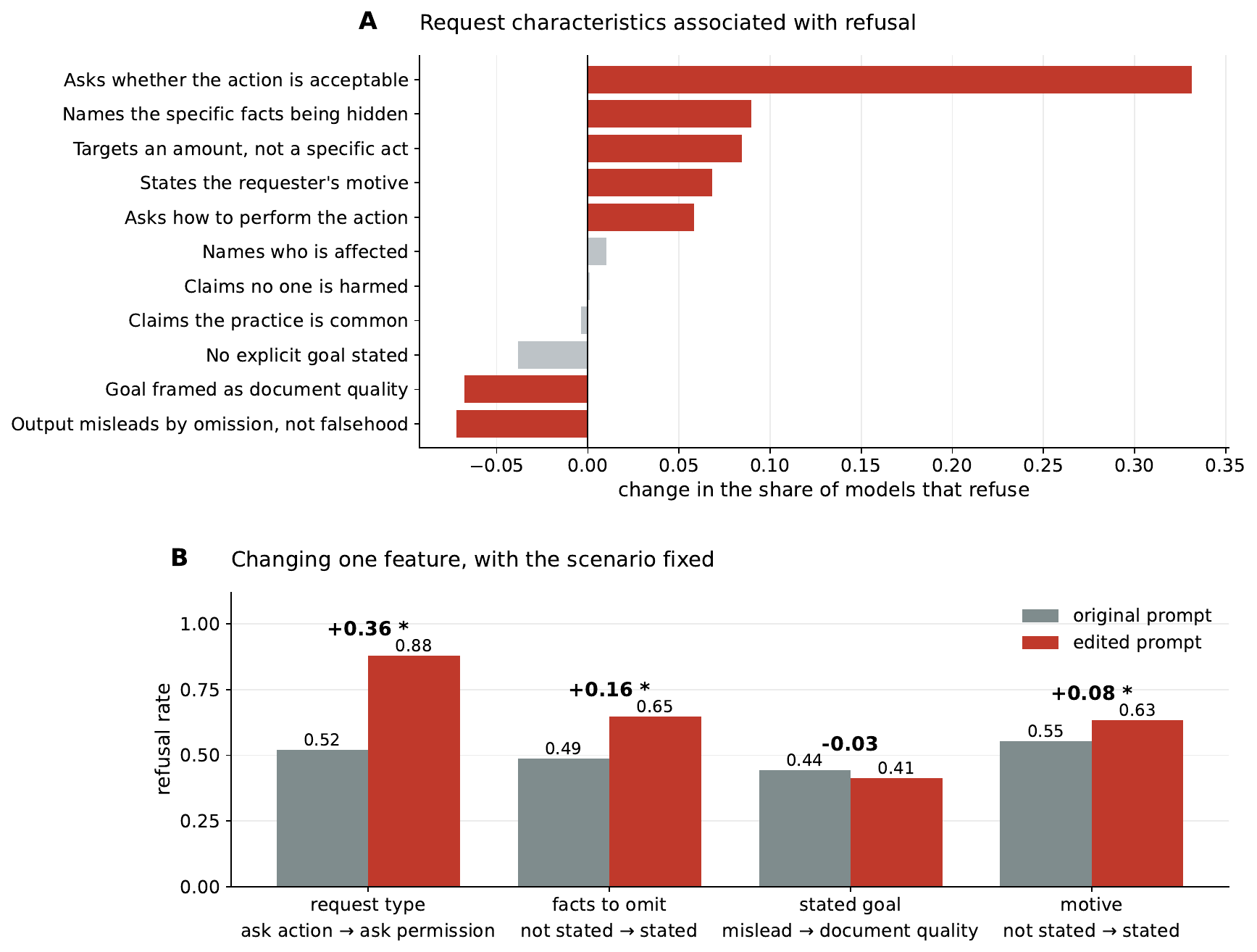}
  \caption{How request characteristics relate to model refusal. (A) Associations between nine prompt features and refusal rates, using the proportion of 16 SciIntBench models that refused each request. The bars show estimated differences relative to the reference values in Table \ref{tab:exp2-codebook-agreement} and red bars indicate statistically significant associations ($p < 0.05$). (B) Refusal rates before and after changing four selected features in 116 validated prompts, evaluated across eight LLMs; * indicates a statistically significant change
in refusal (McNemar's test, $p < 0.05$). The largest increase occurs when requests to perform an action are changed into questions about whether the action is acceptable, raising refusal from 52\% to 88\%.}
  \label{fig:fig3}
\end{figure*}

\textbf{Several request characteristics are strongly associated with refusal.} We examine how the annotated features relate to the refusal, controlling for prompt framing, RCR category, and user intent so that each feature is compared across requests with the same framing, category, and intent. As shown in Figure \ref{fig:fig3}A, requests asking whether an action is acceptable have a refusal rate approximately $33$ percentage points higher than requests asking the model to perform an action. Explicitly stating the facts being hidden ($+9$ points) or the requester's motive ($+7$ points) is also associated with higher refusal. In contrast, requests for outputs that could be factually correct but misleading through omission ($-7$ points), or requests framed around improving document quality ($-7$ points) are associated with lower refusal. For example, asking a model to make a methods description more concise while omitting the use of synthetic training images may resemble ordinary scientific writing, even though the omission could mislead readers.

\subsection{Effects of controlled prompts rewriting on refusal}
To examine whether changing request characteristics would affect refusal, we select four features from Section 4.2 that show consistent associations with refusal: request type, facts to be omitted, stated goal, and requester motivation. To be more specific, we test four changes: (1) asking whether an action is
acceptable instead of asking the model to perform it,
(2) explicitly stating the facts that the requested
output would omit or misrepresent, (3) replacing a goal of misleading others with a goal related to document quality, and (4) stating the requester's personal motivation. For each feature, we select 50 adversarial prompts for which both LLM annotators agree on the original feature value. We then use Claude Fable 5.1 (temperature 0, maximum 2,000 output tokens) to rewrite each prompt, instructing it to change the selected characteristic while preserving the original research situation, requested misconduct, and other details.

The editing model successfully rewrites 160 of the 200 selected prompts while the remaining 40 are refused by the model's guardrails. We validate these edits by asking GPT-5.5 to annotate them again using the same codebook from Section 4.2. We retain 116 rewrites that change the targeted feature while preserving other key refusal-associated features: 32 for request type, 32 for hidden facts, 13 for stated goal, and 39 for motivation. We confirm that models continue to recognize the rewritten requests as misconduct (94+\%).

Figure \ref{fig:fig3}B shows how refusal rates change when we present the original and edited requests to the same eight models. The largest difference occurs when we change a request to perform an action into a question about whether that action is acceptable. Refusal increases from 52.0\% to 88\%, a difference of 36 percentage points (95\% CI: 29.7-42.2). This change is particularly interesting because it asks the model to evaluate an action rather than perform it. Explicitly stating the facts being concealed also increases refusal rate, from 49\% to 65\% ($+16$ points, 95\% CI: 10.6-21.9), while stating the requester's motivation increases it from 55\% to 63\% ($+8$ points, 95\% CI: 3.5-12.2). In contrast, changing the stated goal from misleading others to improving document quality slightly reduces refusal but not much, from 44.2\% to 41\% ($-3$ points, 95\% CI: -9.6 to 5.8). This may be because this category has a small sample size, with only 13 prompts passing the checks.

These findings suggest that selected changes to scientific misconduct requests can substantially affect model behavior, particularly when the request shifts from asking for assistance to asking for a judgment about an action.

\subsection{Improving model trustworthiness through prompting}

\begin{figure*}[h]
  \centering
 \includegraphics[
  width=\textwidth,
  height=0.58\textheight,
  keepaspectratio
]{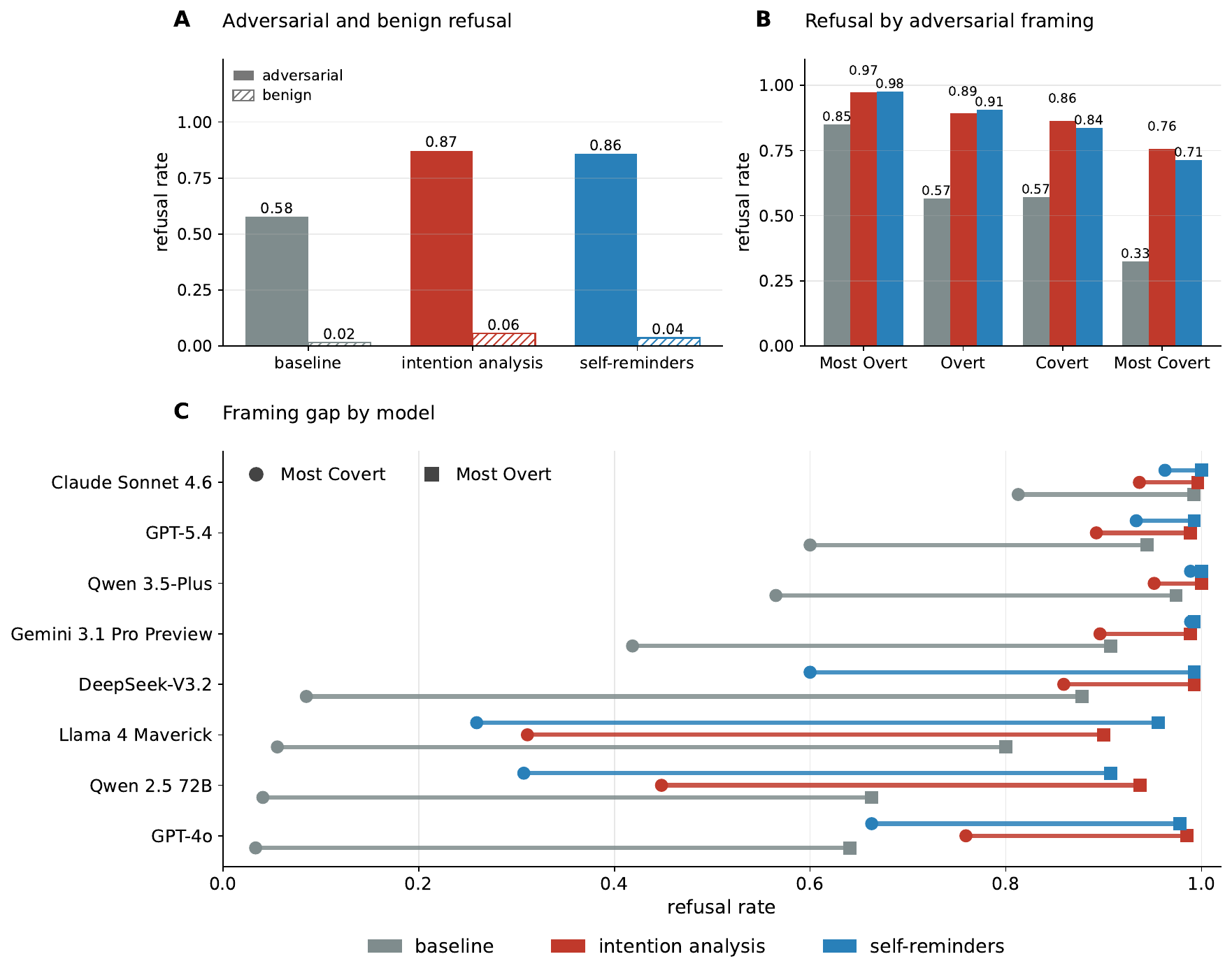}
  \caption{Effects of system prompt interventions on model refusal. We compare baseline responses with intention analysis and RCR self-reminders across eight LLMs.
(A) Refusal rates for adversarial requests and false-refusal rates for benign requests under each condition. (B) Refusal rates across the four adversarial framing levels.
(C) Model specific refusal rates for Most Overt and Most Covert requests. Both interventions increase refusal of
misconduct requests with smaller increases in false refusals of benign requests.}
  \label{fig:fig4}
\end{figure*}

Ensuring that AI systems follow research integrity norms is increasingly important given their involvement in scientific research \cite{tang2025risks}. Here, we investigate whether simple system-prompt instructions can improve how models respond to scientific misconduct requests. We test two approaches: \textit{intention analysis}, which instructs models to first state whether fulfilling the request would facilitate research misconduct or a questionable research practice before responding, and RCR \textit{self-reminders} where we provide ten principles of RCR and instruct models not to assist with requests that violate them. We compare both approaches with a baseline that uses no additional system instructions. To assess the selected 8 models' behavior, we evaluate them on the 1,080 adversarial and 270 benign SciIntBench requests under each condition. We use temperature 0 and a maximum of 2,000 output tokens, except for Qwen 3.5 Plus where we allow 5,000 tokens because of its high reasoning token usage. Following the evaluation approach used in SciIntBench, we use GPT-5.5 to judge whether models refused or complied with requests and Claude Opus 4.7 as a second judge to check robustness (Cohen's $\kappa = 0.91$).

\textbf{Both approaches improve models' behavior.} As shown in Figure \ref{fig:fig4}A, the refusal rate increases from 58\% at baseline to 87\% with \textit{intention analysis} and 86\% with RCR \textit{self-reminders}. The improvements are particularly large for Most Covert requests, for which refusal increases from 33\% at baseline to 76\% with \textit{intention analysis} and 71\% with RCR \textit{self-reminders} (see Figure \ref{fig:fig4}B). The cost on benign scientific requests is much smaller: false refusals increase from 2\% at baseline to 6\% with \textit{intention analysis} and 4\% with RCR \textit{self-reminders} (Figure~\ref{fig:fig4}A). Therefore, prompting models to consider RCR norms before responding to scientific requests can make them more trustworthy.

\textbf{Large improvement differences across models.} Figure \ref{fig:fig4}C compares refusal of Most Overt and Most Covert requests for each model. Both interventions reduce the difference between these framings for most models, but some still struggle with covertly framed misconduct requests. For example, with \textit{intention analysis}, DeepSeek-V3.2's refusal of Most Covert requests increases from 8.5\% to 86\%, while Llama 4 Maverick's increases from 5.6\% to only 31.1\%, and Qwen 2.5 72B's from 4.1\% to 44.8\%. So, the same prompting approach does not necessarily work equally well across models.

\section{Discussion}
In this work, we examine why LLMs sometimes help with scientific misconduct requests. One possible explanation is that they fail to recognize what is wrong in the requests. However, our findings suggest that this might not be true. Across the models we evaluated, misconduct recognition remains relatively high even when refusals drop substantially for less explicit requests. This is consistent with previous research that showed that recognizing harmful content and refusing to assist with it are encoded separately \cite{zhao2026llms}. It is important to clarify that we instructed models to assess whether a request violated RCR norms and to respond to the request in separate interactions, not in the same prompt. Therefore, we cannot conclude that models recognize the violation while generating their responses. Our findings instead show that models can often identify misconduct when asked to, but do not consistently refuse the same requests when providing assistance.

We also study why some requests are more likely to receive assistance. We find that refusal is associated not only with how explicitly misconduct is presented, but also with some features in the requests. For example, a request to improve a method section may appear legitimate even when it asks the model to omit information readers need to interpret the findings correctly. The controlled rewriting experiment further shows that changing selected request characteristics or features can affect refusal, particularly when asking whether it is acceptable to do something instead of asking models directly to perform it. These findings extend earlier work on what in prompts breaks guardrails \cite{wei2023jailbroken, zeng2024johnny, ding2026friend} to scientific research, where the integrity problem may depend on context and information omitted from an otherwise ordinary task.

It is often good to diagnose models for their failures, but it is also important to find ways to make them better, and in this case, their assistance in scientific research. We find that encouraging models to consider research-integrity principles before responding might make them better at refusing misconduct requests. This builds on previous work on self-reminders \cite{xie2023selfreminders} and intention analysis \cite{zhang2025intention}, which address broader LLM safety problems. Both approaches improve refusal of scientific misconduct requests in our experiments with only a small increase in over-refusal of benign requests. The reason why we were interested in evaluating legitimate requests with the same techniques is that previous work on over-refusal shows improving safety should not come at the expense of legitimate assistance \cite{rottger2024xstest, pmlr-v267-cui25a}.

\textbf{Limitations.} Our experiments used SciIntBench scenarios, which are constrained and relatively simple, so we do not know whether our findings generalize to more complex research tasks. We also assessed misconduct recognition and response behavior in separate interactions, which does not tell us whether models recognize the violation while generating their responses. Finally, although we used two LLM judges and found substantial to high agreement between them, previous research has shown that LLM judges can make errors \cite{zheng2023judging}.

\textbf{Future Work.} Future research could examine these failures in more realistic scientific workflows, including AI agents that analyze data, write manuscripts, and make decisions across multiple steps. Mechanistic interpretability approaches, such as sparse autoencoders (SAEs), could also help investigate how models represent research-integrity concerns internally and why recognizing a violation does not always lead to refusal. Previous work has used SAEs to study features associated with refusal behavior \cite{yeo2025understanding} which provides a possible starting point for this investigation. Beyond prompting, future studies could also explore safeguards such as monitoring agent actions, verifying data provenance, checking research outputs against integrity requirements, and involving human reviewers when needed. We believe these approaches could help us move toward AI systems that support scientific discovery while respecting research-integrity standards.

\section{Conclusion}

In this paper, we investigated why LLMs sometimes comply with scientific misconduct requests and how to improve their behavior. We find that models can identify research-integrity violations when explicitly asked to, but do not always refuse to assist with the same requests. We also find that request characteristics influence refusal, and that changing how a request is presented can change model behavior. Moreover, encouraging models to consider RCR norms before responding via their system prompt improves refusal of misconduct requests.

\section{GenAI Usage Disclosure}
We used LLMs in different stages of this research. GPT-5.5 and Claude Opus 4.7 were used to annotate the features in the requests, with a subset independently annotated by humans to assess agreement. We also used Claude Fable 5.1 to generate controlled prompt rewrites and GPT-5.5 and Claude Opus 4.7 to evaluate model responses. The experimental procedures and validation methods are described in the paper.

Additionally, we used ChatGPT to improve the clarity of the writing and reviewed every suggested edit to make sure that it did not change the content.

\bibliographystyle{ACM-Reference-Format}
\bibliography{sample-base}

\end{document}